\documentclass[runningheads]{llncs}
\usepackage{graphicx}
\usepackage[font=small,skip=1pt]{caption}
\usepackage{float}
\usepackage{caption}
\usepackage{subcaption}
\usepackage{hyperref,xcolor}
\begin{document}
\title{BLEU, METEOR, BERTScore: Evaluation of Metrics Performance in Assessing Critical Translation Errors in Sentiment-oriented Text}

\titlerunning{Evaluating Metrics}
%
\author{Hadeel Saadany\inst{1} \and
Constantin Or\u{a}san\inst{2}}
\authorrunning{Saadany et al.}
%
\institute{University of Wolverhampton, UK, 
\email{h.a.saadany@wlv.ac.uk}\and
University of Surrey, UK, \email{c.orasan@surrey.ac.uk}
\\
}
\maketitle              

\begin{abstract}
Social media companies as well as authorities make extensive use of artificial intelligence (AI) tools to monitor postings of hate speech, celebrations of violence or profanity. Since AI software
requires massive volumes of data to train computers, Machine Translation (MT) of the online content is commonly used to process posts written in several languages and hence augment the data needed for training. However, MT mistakes are a regular occurrence when translating sentiment-oriented user-generated content (UGC), especially when a low-resource language is involved. The adequacy of the whole process relies on the assumption that  the evaluation metrics used give a reliable indication of the quality of the translation. In this paper, we assess the ability of automatic quality metrics to detect critical machine translation errors which can cause serious misunderstanding of the affect message. We compare the performance of three canonical metrics 
on meaningless translations where the semantic content is seriously impaired as compared to meaningful translations with a critical error which exclusively distorts the sentiment of the source text. We conclude that there is a need for fine-tuning of automatic metrics to make them more robust in detecting sentiment critical errors.

\end{abstract}

\section{Introduction}

Facebook has once apologised after its machine-translation service lead to an arrest of a man from the West Bank whose profile posting in his native dialect that read ``good morning''  was mistranslated as ``attack them'', and later  automatically detected by authorities as an incitement to violence\footnote{\url{https://www.theguardian.com/technology/2017/oct/24/facebook-palestine-israel-translates-good-morning-attack-them-arrest}}. The main danger in this type of MT error is that it changes the author's sentiment, here from positive to a negative or rather aggressive emotion. Research on translation of sentiment by MT systems has shown that users encounter similar mistakes where the sentiment polarity of the source is flipped to its exact opposite due to a mistranslation of a contronym, a dialectical expression, or a missed negation marker, especially in translation of online content of low-resource languages \cite{saadany2020great}. In machine translation research, the reliability of MT systems is conventionally measured by automatic quality metrics such as BLEU \cite{papineni2002bleu} and METEOR \cite{banerjee2005meteor}. The aim of these automatic quality metrics is to evaluate a translation hypothesis (i.e. the automatic translation) against a reference translation, which is normally produced by a human translator. Good evaluation metrics should have a high correlation with human judgement on the quality of translation. Recently some automatic metrics have achieved a significant correlation with human judgement on the WMT Metrics task datasets (see \cite{openkiwi,yisi20,2020mee}). However, research has reported weaker correlation with low human assessment score ranges for segment-level evaluation \cite{takahashi2020automatic,takashi21translation}. These findings point to the challenges involved in detecting low-quality translations by automatic metrics.

In this work, we focus on the problem of evaluating critical translation errors that can cause serious misunderstanding of the sentiment conveyed in the source text. To illustrate this point, suppose we are evaluating the MT output ``\textit{People are dead, starving in your presence, may God forgive you}" with its reference ``\textit{People are dead, starving in your presence, may God \textbf{not} forgive you}"\footnote{The hypothesis is the mistranslation of Twitter's Translate tab for an Arabic tweet \url{https://twitter.com/ZPNyOawCRVTNBxu/status/878496659793170432}, accessed 26 June 2021.}. The error in the MT output is only the missing of the word \textit{not}, however, this omission causes the translation to convey the exact opposite sentiment of the source. We argue that such translation errors should be considered more critical than those which produce ungrammatical or low-quality translations, but do not significantly distort the message of the source. However, as we show in this paper, automatic quality metrics fail to give a penalty to this type of critical error proportional to its gravity and may equate this hypothesis with another that also has a uni-gram mistake, but transfers the affect message (e.g \textit{People are dead, \textbf{hungry} in your presence, may God not forgive you}). 


In this research we conduct an experiment with three canonical automatic quality metrics, BLEU, METEOR and BERTScore \cite{bertscore}. We measure the ability of each metric to penalise sentiment critical errors that severely distort the affect message as compared to translations which correctly transfer the correct sentiment as well as mistranslations that produce incomprehensible content in the target language. We first briefly present the three metrics in section \ref{sec1}. Then, in section \ref{sec2}, we explain our experiment and summarise the results. In section \ref{sec3}, we give our concluding remarks.
\vspace{-.3cm}
\section{Related Work}
\label{sec1}
The standard metric for assessing empirical improvement of MT systems is BLEU. Simply stated, the objective of BLEU is to compare n-grams of the candidate translation with n-grams of the reference translation and count the number of matches; the more the matches, the better the candidate translation. The final score is a modified n-gram precision multiplied by a brevity penalty to account for both frequency and adequacy.  Due to its restrictive exact matching to the reference, BLEU does not accommodate for importance n-gram weighting which may be essential in assessing a sentiment-critical error. However, despite research evidence of  its analytical limitations \cite{tangledbleu,structuredrviewbleu}, BLEU, is still the \textit{de facto} standard for MT performance evaluation  because it is easy to calculate regardless of the languages involved. METEOR, on the other hand, incorporates semantic information as it evaluates translation by calculating either exact match, stem match, or synonymy match. For synonym matching, it utilises WordNet synsets \cite{wordnet}. More recent versions (METEOR 1.5 and METEOR++2.0) apply importance weighting by giving smaller weight to function words \cite{denkowski2014meteor,guo2019meteor++}. However, the METEOR weighting scheme would not allow for a great penalty of the missing negation marker in the hypothesis of our example above. In fact, the METEOR score for Twitter's MT wrong translation is 0.91, whereas the score for the correct translation (\textit{People are dead, starving in your presence, may God \textbf{not} forgive you}) is 0.99. The main culprit for this proportionally inaccurate scoring is the function word weighting which causes the metric to be over permissive despite the MT engine missing of a negation marker crucial to the sentiment of the source tweet. 

Both METEOR and BLEU assess the quality of translation in terms of surface n-gram matching between the MT output and a human reference(s). After the introduction of pretrained contextual word models, there has been a recent trend to use large-scale models like BERT  \cite{bert} for MT evaluation to incorporate semantic contextual information of tokens in comparing translation and reference segments.  A number of embedding-based metrics has proven to achieve the highest performance in recent WMT shared tasks for quality metrics (e.g. \cite{openkiwi,yisi20,2020mee}). We take BERTScore as representative of this category. BERTScore computes a score based on a pair wise cosine similarity between the BERT contextual embeddings of the individual tokens for the hypothesis and the reference. Accordingly, a BERTScore close to 1 indicates proximity in vector space and hence a good translation. In the following section, we explain our experiment for assessing the performance of these three metrics with respect to critical translation errors that seriously distort the affect message of the source.

\section{Experiment Set Up}
\label{sec2}
\subsection{Dataset Compiling}

We measure the performance of the three metrics on two types of translated UGC data: synthetic and authentic. The synthetic dataset consists of 100 restaurant reviews extracted from the SemEval-2016 Aspect-Based Sentiment Analysis task where each review expresses mixed sentiment about a particular entity \cite{ASPECT}. For this dataset we did not use machine translation, but we artificially modified the original texts in such a way that the original sentiment was distorted. Thus, we created hypothesis-reference pairs with changes only in sentiment-related words. The main objective of the synthetic data is to measure the sensitivity of each metric to sentiment-critical translation errors by making n-gram sentiment modifications to the hypothesis while keeping the other words intact. We made four types of sentiment modifications manually. For example, for the source review `\textit{But the staff was so horrible to us}', we made the following modifications:

\begin{itemize}

    \item One Non-Critical Error: a uni-gram change that does not affect the sentiment (`\textit{But the staff was so horrible to \textbf{him}}')
    
    \item One Critical Error: a uni-gram change that produced the opposite sentiment (`\textit{But the staff was so \textbf{nice} to us}')
    \item Two Errors: a two-words change with one critical and one non-critical error (`\textit{But the staff was so \textbf{nice} to \textbf{him}}')
    \item Nonsense: a three-words change that produced a meaningless translation (`\textit{But the team was so to him}')

\end{itemize}
\vspace{-.9cm}

\begin{table}[htp]

\caption{Distribution of Translation of Sentiment Errors for the Datasets}\label{tab1}
\centering
\begin{tabular}{l|c|c|c|c}
\hline
\textbf{Dataset} &
  \textbf{\ No Error\ } &
  \textbf{\ One Error\ } &
  \textbf{\ Two Errors\ } &
  \textbf{\ Nonsense\ } \\ \hline
\textbf{Synthetic En to En}             &         & 200    & 100    & 100    \\ \hline
\textbf{Total}                 & \multicolumn{4}{c}{\textbf{400}}  \\ \hline
\textbf{\begin{tabular}[c]{@{}l@{}}Authentic En to\\ Sp/Ar/Pt/Ro\end{tabular}} &
  854 &
  404 &
  142 &
   \\ \hline
\textbf{Authentic Sp/Ar to En\ } & 150     & 150    &        &        \\ \hline
\textbf{Total}                 & \multicolumn{4}{c}{\textbf{1700}} \\ \hline
\end{tabular}
\end{table}

The authentic dataset consisted of 1700 tweets collated from different emotion-detection and aggression-detection shared tasks (\cite{shared1,shared2,shared3,shared4}). The source tweets were in three languages: English (1400), Arabic (200) and Spanish (100). This dataset was translated by Twitter's MT system (Google API). The Spanish and English source tweets were translated into English, and the English tweets were translated into Romanian, Arabic, Spanish and Portuguese. Five human annotators \footnote{The annotators were computational linguists working on MT research.}, native speakers of the respective languages, manually annotated the translations for sentiment errors. The annotation was straightforward: \textit{Yes} the translation transfers the sentiment of the source (even though it can have non-sentiment related errors that do not seriously affect the overall sentiment/emotion) or \textit{No}, it does not. If `No', the annotators were asked to mark whether the mistranslation of sentiment is due to one or two linguistic errors. The linguistic error was either a missing negation marker, a mistranslation of a hashtag, an idiomatic expression or a polysemous word (table \ref{tab1} shows the distribution of the datasets types used in the experiment). More details on how the errors were identified are discussed in \cite{saadany2021challenges}. 

We ran the three metrics on the hypothesis/reference pairs of the synthetic dataset and the hypothesis/reference\footnote{Reference translations were created by the two annotators native speakers of Arabic and Spanish.} for Arabic and Spanish tweets, and the source/back-translations of the English tweets of the authentic dataset (The back-translations were checked to make sure they reproduced the exact sentiment errors in the MT output). Accordingly, as shown in table \ref{tab1}, we evaluated 400 synthetic English hypothesis/reference pairs, 1400 English tweets translated into Romanian, Arabic, Spanish and Portuguese, and 300 Arabic and Spanish tweets translated into English. We used these datasets to calculate three measures for BLEU, METEOR and BERTScore: segment-level scores, mean segment-level scores and standard deviation for segment-level scores. Results of the experiment are explained in the next section.




\begin{figure}[tp]
\centering
\begin{minipage}{.53\textwidth}
  \centering
  \includegraphics[scale=.48,trim={0cm 0cm 0cm .0cm},clip]{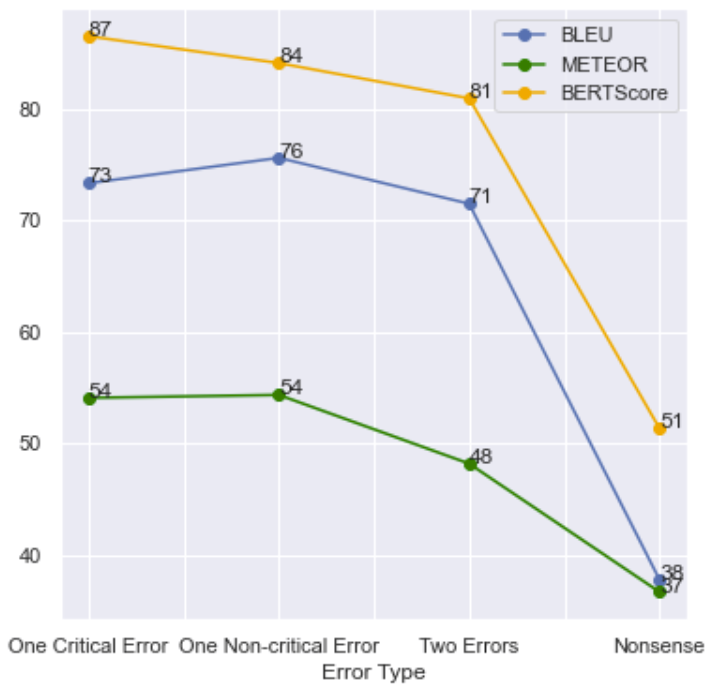}
  \caption[Caption for LOF]{Mean Scores for Synthetic Data\protect\footnotemark}
  \label{fig:test1}
\end{minipage}%
\begin{minipage}{.5\textwidth}
  \centering
  \includegraphics[scale=.48,trim={0cm 0cm 0cm 0cm},clip]{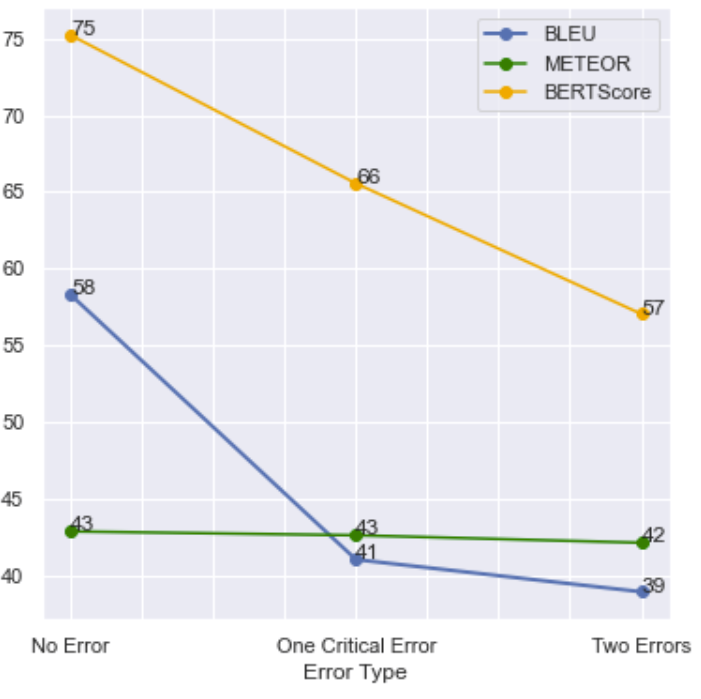}
  \captionof{figure}{Mean Scores for Authentic Data (en)}
  \label{fig:test2}
\end{minipage}
\begin{minipage}{.5\textwidth}
  \centering
  \includegraphics[scale=.48,trim={0cm 0cm 0cm 0cm},clip]{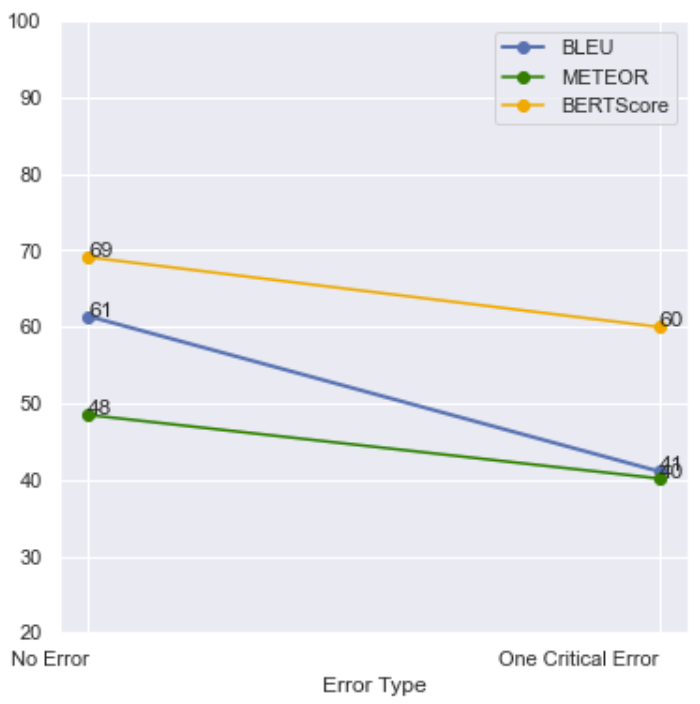}
  \captionof{figure}{Mean Scores for Authentic Data (ar/sp)}
  \label{fig:test3}
\end{minipage}%
\begin{minipage}{.5\textwidth}
  \centering
  \includegraphics[scale=.47,trim={0cm 0cm 0cm 0cm},clip]{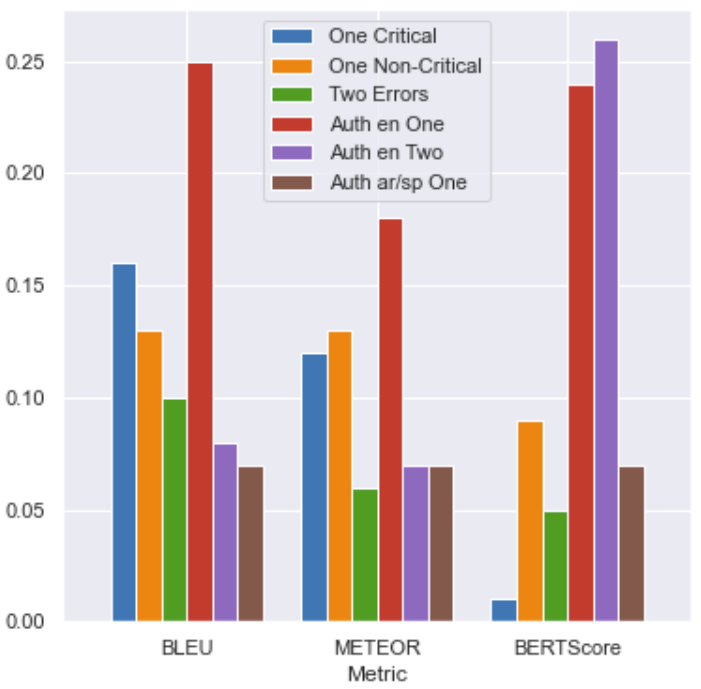}
  \captionof{figure}{Normalised Standard Deviation}
  \label{fig:test4}
\end{minipage}
\end{figure}
\footnotetext{Scores in figures are standardised from 0 to 100 for easier display.}

\subsection{Results}

The average segment scores of the three metrics for the four sentiment modifications we have conducted on the hypotheses of the synthetic dataset is shown in figure \ref{fig:test1}.
As can be seen from the figure, the difference between the mean score for one critical error and one non-critical error is quite small for all the three metrics (max 3 points difference). This result essentially highlights the inability of the three metrics to distinguish between the mistranslation of a critical word that seriously distorts the affect message and the mistranslation of a non-critical word that does not affect the sentiment content (see table \ref{tab2} for examples of such cases). The metrics, however, are able to distinguish low-quality translation with a highly distorted content as the average scores for the `Nonsense' translations are far off from the other types of errors.  Furthermore, the average BLEU score for one non-critical error is slightly higher than the one critical error. This is due to the fact that BLEU gauges the performance of an MT model by an indiscriminate n-gram matching, regardless of the semantic weight of each word. An error with a sentiment-critical word, therefore, is equally penalised as any other word. Also, for BERTScore the average score for one critical error is relatively high (0.85) due to what is known as the antonymy problem in contextual word embeddings \cite{antonymy}. Antonyms (e.g. `great' and `terrible') usually have similar contextual information and hence are closer in vector space. The change of one word to its exact opposite, therefore, is not adequately captured by the BERTScore metric. It can be claimed, therefore, that the embedding-based metric would generally struggle with hypotheses with only a uni-gram sentiment-critical error that flips the source sentiment to its opposite polarity. 

Figure \ref{fig:test2} shows a similar problem for the authentic English data. For METEOR, a translation that transfers the affect message has a similar average score as translations that have one or two linguistic errors that seriously distort the sentiment of the source. Note that in the authentic `No Error' dataset, the hypothesis correctly transfers the main content but may have non-sentiment errors and hence METEOR scores may be lower for some hypotheses. However, the METEOR performance casts doubt on its ability to distinguish between a translation that can transmit the sentiment content despite other errors and another translation that has a critical error of the sentiment which would be unacceptable by human standards. By contrast, the average scores of the BERTScore metric correlate consistently with the degradation of the sentiment transfer in this authentic dataset. However, for the second language arc where Arabic/Spanish are the source languages, the difference between METEOR and BERTScore average scores for segments with no sentiment error and those with critical errors is relatively small (7 and 8 points, respectively as shown in figure\ref{fig:test3}).  

 Finally, figure \ref{fig:test4} shows the normalised standard deviation of the segment-level scores for the three metrics on the different datasets. The scores of the three metrics display the highest variation with the authentic dataset with one sentiment error and BERTScore displays a great variance with two sentiment errors in the same dataset. This indicates that translations with sentiment critical errors do not consistently receive low scores by the three metrics. Similarly, both the METEOR and BLEU metrics have a relatively higher deviation in segment-level scores for the synthetic dataset with one critical error. Therefore, hypotheses that are exact match to the reference but have only one critical error causing a misinterpretation of the affect message are not consistently penalised by the two metrics (see table \ref{tab2} for examples of metric scores for references/hypotheses of the two datasets). 
\vspace*{-\baselineskip}
\vspace*{-\baselineskip}

\begin{table}[htp]
\caption{Examples of Metric Scores for Different Error Types}
\label{tab2}
\begin{tabular}{|l|l|l|l|l|}

\hline
\multicolumn{2}{|l|}{Synthetic Data}                                                                & \multicolumn{3}{l|}{Metric}                 \\ \hline
\multicolumn{2}{|l|}{}                                                                                       & BLEU & METEOR & BERTScore \\ \hline
Ref       & \begin{tabular}[c]{@{}l@{}}Their pizza is the best, \\ if you like thin crusted pizza.\end{tabular} & 1.0           & 1.0             & 1.0                \\ \hline
Non-critical Error & \begin{tabular}[c]{@{}l@{}}Their pizza is the best, \\ if you like thin \textbf{\textit{layer}} pizza.\end{tabular}    & 0.76 & 0.50 & 0.90 \\ \hline
Critical Error     & \begin{tabular}[c]{@{}l@{}}Their pizza is the \textbf{\textit{worst}}, \\ if you like thin crusted pizza.\end{tabular} & 0.73 & 0.50 & 0.86 \\ \hline
\multicolumn{2}{|l|}{Authentic Data}                                                                &               &                 &                    \\ \hline
Ref       & \begin{tabular}[c]{@{}l@{}}What is this amount of happiness,\\  I don't understand!\end{tabular} & 1.0           & 1.0             & 1.0                \\ \hline
One Error & \begin{tabular}[c]{@{}l@{}}What is this amount of \textbf{\textit{anger}},\\ I don't get it!\end{tabular}          & 0.65          & 0.47            & 0.89               \\ \hline
Ref       & \begin{tabular}[c]{@{}l@{}}Sweetie like clouds,\\  always fill me with joy.\end{tabular}          & 1.0           & 1.0             & 1.0                \\ \hline
No Error  & \begin{tabular}[c]{@{}l@{}}\textbf{\textit{My love}} is like clouds,\\  always fill me with joy.\end{tabular}       & 0.65          & 0.44            & 0.52               \\ \hline
\end{tabular}
\end{table}

\vspace*{-\baselineskip}
\vspace*{-\baselineskip}
\section{Conclusion}
\label{sec3}

In this research, we conducted an experiment with three canonical automatic quality metrics to evaluate their ability to penalise a critical translation error that seriously distorts the affect message of the source text. The average segment-level scores for the three metrics showed that sentiment-critical and non-critical errors are not appropriately distinguishable especially in our synthetic dataset. This shows that in scenarios  where the MT output is an exact match to the reference except for one sentiment-pivotal word, the automatic quality metric becomes less sensitive to the mistranslation error. Similarly, with the authentic datasets, the average scores for METEOR showed that mistranslations with one or two critical errors are not appropriately penalised. Moreover, with both the authentic and synthetic data, the relatively high inconsistency of segment-level scores for hypotheses with one or two sentiment-critical errors suggests that a distortion of the sentiment content may misleadingly receive high scores by any of the three metrics. The results of the experiment call attention to the need for a sentiment-targeted evaluation measure that can adequately assess this type of critical translation errors that have can serious consequences in determining the sentiment stance of the author. Our future work will focus on fine-tuning the quality metrics to capture sentiment-critical lexicon to improve its performance with sentiment-oriented text.

\section*{Acknowledgements}
Part of the research done by Hadeel Saadany was carried out in the context of the TranSent project at the University of Surrey. 

\bibliographystyle{splncs04}
\bibliography{ranlp2021}

\begin{thebibliography}{10}
\providecommand{\url}[1]{\texttt{#1}}
\providecommand{\urlprefix}{URL }
\providecommand{\doi}[1]{https://doi.org/#1}

\bibitem{banerjee2005meteor}
Banerjee, S., Lavie, A.: {METEOR: An automatic metric for MT evaluation with
  improved correlation with human judgments}. In: Proceedings of the acl
  workshop on intrinsic and extrinsic evaluation measures for machine
  translation and/or summarization. pp. 65--72 (2005)

\bibitem{shared3}
Basile, V., Bosco, C., Fersini, E., Debora, N., Patti, V., Pardo, F.M.R.,
  Rosso, P., Sanguinetti, M., et~al.: Semeval-2019 task 5: Multilingual
  detection of hate speech against immigrants and women in twitter. In: 13th
  International Workshop on Semantic Evaluation. pp. 54--63. Association for
  Computational Linguistics (2019)

\bibitem{denkowski2014meteor}
Denkowski, M., Lavie, A.: Meteor universal: Language specific translation
  evaluation for any target language. In: Proceedings of the ninth workshop on
  statistical machine translation. pp. 376--380 (2014)

\bibitem{bert}
Devlin, J., Chang, M.W., Lee, K., Toutanova, K.: Bert: Pre-training of deep
  bidirectional transformers for language understanding. arXiv preprint
  arXiv:1810.04805  (2018)

\bibitem{antonymy}
Etcheverry, M., Wonsever, D.: Unraveling antonym’s word vectors through a
  siamese-like network. In: Proceedings of the 57th Annual Meeting of the
  Association for Computational Linguistics. pp. 3297--3307 (2019)

\bibitem{guo2019meteor++}
Guo, Y., Hu, J.: Meteor++ 2.0: Adopt syntactic level paraphrase knowledge into
  machine translation evaluation. In: Proceedings of the Fourth Conference on
  Machine Translation (Volume 2: Shared Task Papers, Day 1). pp. 501--506
  (2019)

\bibitem{openkiwi}
Kepler, F., Tr{\'e}nous, J., Treviso, M., Vera, M., Martins, A.F.: Openkiwi: An
  open source framework for quality estimation. arXiv preprint arXiv:1902.08646
   (2019)

\bibitem{yisi20}
Lo, C.k.: {Extended study on using pretrained language models and YiSi-1 for
  machine translation evaluation}. In: Proceedings of the Fifth Conference on
  Machine Translation. pp. 895--902 (2020)

\bibitem{tangledbleu}
Mathur, N., Baldwin, T., Cohn, T.: {Tangled up in BLEU: Reevaluating the
  Evaluation of Automatic Machine Translation Evaluation Metrics}. arXiv
  preprint arXiv:2006.06264  (2020)

\bibitem{shared2}
Mohammad, S., Kiritchenko, S.: Understanding emotions: A dataset of tweets to
  study interactions between affect categories. In: Proceedings of the Eleventh
  International Conference on Language Resources and Evaluation (LREC 2018)
  (2018)

\bibitem{shared1}
Mohammad, S.M., Bravo-Marquez, F.: Wassa-2017 shared task on emotion intensity.
  arXiv preprint arXiv:1708.03700  (2017)

\bibitem{2020mee}
Mukherjee, A., Ala, H., Shrivastava, M., Sharma, D.M.: Mee: An automatic metric
  for evaluation using embeddings for machine translation. In: 2020 IEEE 7th
  International Conference on Data Science and Advanced Analytics (DSAA). pp.
  292--299. IEEE (2020)

\bibitem{papineni2002bleu}
Papineni, K., Roukos, S., Ward, T., Zhu, W.J.: Bleu: a method for automatic
  evaluation of machine translation. In: Proceedings of the 40th annual meeting
  of the Association for Computational Linguistics. pp. 311--318 (2002)

\bibitem{wordnet}
Pedersen, T., Patwardhan, S., Michelizzi, J., et~al.: Wordnet::
  Similarity-measuring the relatedness of concepts. In: AAAI. vol.~4, pp.
  25--29 (2004)

\bibitem{ASPECT}
Pontiki, M., Galanis, D., Papageorgiou, H., Androutsopoulos, I., Manandhar, S.,
  Al-Smadi, M., Al-Ayyoub, M., Zhao, Y., Qin, B., De~Clercq, O., et~al.:
  Semeval-2016 task 5: Aspect based sentiment analysis. In: International
  workshop on semantic evaluation. pp. 19--30 (2016)

\bibitem{structuredrviewbleu}
Reiter, E.: {A Structured Review of the Validity of BLEU}. Computational
  Linguistics  \textbf{44}(3),  393--401 (2018)

\bibitem{saadany2020great}
Saadany, H., Orasan, C.: Is it great or terrible? preserving sentiment in
  neural machine translation of arabic reviews. In: Proceedings of the Fifth
  Arabic Natural Language Processing Workshop. pp. 24--37 (2020)

\bibitem{saadany2021challenges}
Saadany, H., Orasan, C., Quintana, R.C., do~Carmo, F., Zilio, L.: {Challenges
  in Translation of Emotions in Multilingual User-Generated Content: Twitter as
  a Case Study}. arXiv preprint arXiv:2106.10719  (2021)

\bibitem{takashi21translation}
Sudoh, K., Takahashi, K., Nakamura, S.: Is this translation error critical?:
  Classification-based human and automatic machine translation evaluation
  focusing on critical errors. In: Proceedings of the Workshop on Human
  Evaluation of NLP Systems (HumEval). pp. 46--55 (2021)

\bibitem{takahashi2020automatic}
Takahashi, K., Sudoh, K., Nakamura, S.: Automatic machine translation
  evaluation using source language inputs and cross-lingual language model. In:
  Proceedings of the 58th Annual Meeting of the Association for Computational
  Linguistics. pp. 3553--3558 (2020)

\bibitem{shared4}
Zampieri, M., Nakov, P., Rosenthal, S., Atanasova, P., Karadzhov, G., Mubarak,
  H., Derczynski, L., Pitenis, Z., {\c{C}}{\"o}ltekin, {\c{C}}.: Semeval-2020
  task 12: Multilingual offensive language identification in social media
  (offenseval 2020). arXiv preprint arXiv:2006.07235  (2020)

\bibitem{bertscore}
Zhang, T., Kishore, V., Wu, F., Weinberger, K.Q., Artzi, Y.: {Bertscore:
  Evaluating text generation with Bert}. arXiv preprint arXiv:1904.09675
  (2019)

\end{thebibliography}





\end{document}